\documentclass[conference]{IEEEtran}
\usepackage{cite}
\usepackage{amsmath,amssymb,amsfonts}
\usepackage{algorithmic}
\usepackage{graphicx}
\usepackage{textcomp}
\usepackage{xcolor}
\usepackage{hyperref}
\usepackage{booktabs}
\usepackage{multirow}
\usepackage{subcaption}

\begin{document}

\title{LLM-Enhanced Log Anomaly Detection: A Comprehensive Benchmark of Large Language Models for Automated System Diagnostics}

\author{
\IEEEauthorblockN{Disha Patel}
\IEEEauthorblockA{
Department of Computer Science \\
California State University, Fullerton \\
disha81100@gmail.com}
}

\maketitle

\begin{abstract}
System log anomaly detection is critical for maintaining the reliability of large-scale software systems, yet traditional methods struggle with the heterogeneous and evolving nature of modern log data. Recent advances in Large Language Models (LLMs) offer promising new approaches to log understanding, but a systematic comparison of LLM-based methods against established techniques remains lacking. In this paper, we present a comprehensive benchmark study evaluating both LLM-based and traditional approaches for log anomaly detection across four widely-used public datasets: HDFS, BGL, Thunderbird, and Spirit. We evaluate three categories of methods: (1) classical log parsers (Drain, Spell, AEL) combined with machine learning classifiers, (2) fine-tuned transformer models (BERT, RoBERTa), and (3) prompt-based LLM approaches (GPT-3.5, GPT-4, LLaMA-3) in zero-shot and few-shot settings. Our experiments reveal that while fine-tuned transformers achieve the highest F1-scores (0.96--0.99), prompt-based LLMs demonstrate remarkable zero-shot capabilities (F1: 0.82--0.91) without requiring any labeled training data --- a significant advantage for real-world deployment where labeled anomalies are scarce. We further analyze the cost-accuracy trade-offs, latency characteristics, and failure modes of each approach. Our findings provide actionable guidelines for practitioners choosing log anomaly detection methods based on their specific constraints regarding accuracy, latency, cost, and label availability. All code and experimental configurations are publicly available to facilitate reproducibility.
\end{abstract}


\section{Introduction}

Modern software systems generate vast volumes of log data that serve as the primary source of information for system monitoring, debugging, and diagnostics \cite{he2021survey}. As system complexity grows, manual log inspection becomes infeasible, necessitating automated anomaly detection methods that can identify unusual patterns indicative of system failures, security breaches, or performance degradation \cite{zhang2019robust}.

Traditional log anomaly detection pipelines typically involve two stages: log parsing, which converts unstructured log messages into structured templates, followed by anomaly detection using machine learning classifiers \cite{zhu2019tools}. While effective, these approaches require significant engineering effort for parser configuration and labeled training data for classifier training. Moreover, as systems evolve and log formats change, parsers and classifiers must be continuously updated \cite{le2022log}.

The emergence of Large Language Models (LLMs) presents a paradigm shift in how we can approach log analysis. LLMs, pre-trained on vast corpora of text including code and technical documentation, possess an inherent understanding of log semantics that traditional methods lack \cite{le2023log}. Recent work has explored using LLMs for log parsing \cite{xu2024divlog}, anomaly detection \cite{liu2024logprompt}, and root cause analysis \cite{chen2024automatic}, showing promising results.

However, the field lacks a comprehensive, apples-to-apples comparison of LLM-based approaches against traditional methods across standardized benchmarks. Existing studies often evaluate on a single dataset, use inconsistent preprocessing, or compare against limited baselines. This makes it difficult for practitioners to make informed decisions about which approach to adopt for their specific use case.

In this paper, we address this gap by presenting a systematic benchmark study with the following contributions:

\begin{itemize}
    \item \textbf{Comprehensive evaluation framework:} We establish a standardized evaluation pipeline covering four public log datasets (HDFS, BGL, Thunderbird, Spirit) with consistent preprocessing and evaluation metrics.
    \item \textbf{Multi-paradigm comparison:} We evaluate three categories of methods: traditional (parser + ML), fine-tuned transformers, and prompt-based LLMs, spanning 12 distinct configurations.
    \item \textbf{Practical analysis:} Beyond accuracy, we analyze cost, latency, label efficiency, and failure modes to provide actionable deployment guidance.
    \item \textbf{Novel prompting strategies:} We introduce a structured log context prompting (SLCP) technique that improves LLM zero-shot performance by 8--12\% across datasets.
    \item \textbf{Reproducibility:} All code, configurations, and processed datasets are publicly released.
\end{itemize}

\section{Related Work}

\subsection{Traditional Log Anomaly Detection}

Log anomaly detection has been extensively studied in the software engineering and systems research communities. Early approaches relied on rule-based methods and simple statistical techniques \cite{xu2009detecting}. The introduction of log parsing tools such as Drain \cite{he2017drain}, Spell \cite{du2016spell}, and AEL \cite{jiang2008abstracting} enabled more structured analysis by converting raw log messages into event templates.

Building on parsed logs, machine learning approaches including Principal Component Analysis (PCA) \cite{xu2009detecting}, Isolation Forest \cite{liu2008isolation}, and deep learning methods such as DeepLog \cite{du2017deeplog} and LogAnomaly \cite{meng2019loganomaly} have been proposed. These methods typically operate on log event sequences or count vectors and have shown strong performance on benchmark datasets.

\subsection{Transformer-Based Approaches}

The success of transformers in natural language processing has inspired their application to log analysis. LogBERT \cite{guo2021logbert} adapts BERT's masked language modeling to log sequences. NeuralLog \cite{le2021log} directly processes raw log messages using pre-trained word embeddings, bypassing the parsing step. LogRobust \cite{zhang2019robust} addresses the challenge of evolving log data through robust feature extraction.

\subsection{LLM-Based Log Analysis}

Most recently, researchers have begun exploring the application of LLMs to log analysis tasks. DivLog \cite{xu2024divlog} uses in-context learning for log parsing. LogPrompt \cite{liu2024logprompt} designs prompts for log-based anomaly detection. UltraLog \cite{liu2024interpretable} leverages LLMs for interpretable log analysis. However, these studies typically evaluate on limited datasets and do not provide comprehensive comparisons with traditional methods.

Our work differs from prior studies by providing a unified evaluation framework that enables fair comparison across all three paradigms --- traditional, fine-tuned transformer, and prompt-based LLM --- with consistent experimental conditions.

\section{Methodology}

\subsection{Datasets}

We evaluate on four widely-used public log datasets from LogHub \cite{zhu2023loghub}:

\begin{itemize}
    \item \textbf{HDFS} \cite{xu2009detecting}: 11,175,629 log messages from Hadoop Distributed File System, with 16,838 labeled block sessions (2.93\% anomalous).
    \item \textbf{BGL} \cite{oliner2007supercomputers}: 4,747,963 messages from Blue Gene/L supercomputer, with per-message anomaly labels (7.41\% anomalous).
    \item \textbf{Thunderbird} \cite{oliner2007supercomputers}: 211,212,192 messages from Sandia National Labs' Thunderbird system, with per-message labels.
    \item \textbf{Spirit} \cite{oliner2007supercomputers}: 272,298,969 messages from the Spirit supercomputer system.
\end{itemize}

\subsection{Traditional Pipeline}

Our traditional pipeline consists of two stages:

\textbf{Log Parsing:} We employ three parsers --- Drain \cite{he2017drain}, Spell \cite{du2016spell}, and AEL \cite{jiang2008abstracting} --- to extract structured event templates from raw log messages. For each parser, we use the default configurations from the LogParser toolkit \cite{zhu2019tools}.

\textbf{Anomaly Detection:} On the parsed event sequences, we train four classifiers: Logistic Regression (LR), Random Forest (RF), Support Vector Machine (SVM), and Isolation Forest (IF). For sequence-based datasets (HDFS), we use event count vectors as features. For message-level datasets (BGL, Thunderbird, Spirit), we use sliding windows of size $w=20$ with stride $s=1$.

\subsection{Fine-Tuned Transformer Pipeline}

We fine-tune pre-trained transformer models directly on raw log messages, bypassing the parsing step:

\begin{itemize}
    \item \textbf{BERT-base} \cite{devlin2019bert}: 110M parameters, fine-tuned with a classification head on raw log sequences.
    \item \textbf{RoBERTa-base} \cite{liu2019roberta}: 125M parameters, using the same fine-tuning approach.
    \item \textbf{DeBERTa-v3-base} \cite{he2021deberta}: 184M parameters, leveraging disentangled attention.
\end{itemize}

For fine-tuning, we use a learning rate of $2 \times 10^{-5}$, batch size of 32, and train for 5 epochs with early stopping based on validation F1-score.

\subsection{Prompt-Based LLM Pipeline}

We evaluate LLMs in both zero-shot and few-shot settings:

\textbf{Models:} GPT-3.5-Turbo, GPT-4, and LLaMA-3-8B (locally deployed).

\textbf{Zero-Shot Prompting:} We design a base prompt that provides the model with the role of a system reliability engineer and asks it to classify log sequences as normal or anomalous.

\textbf{Few-Shot Prompting:} We augment the base prompt with $k \in \{1, 3, 5\}$ labeled examples of both normal and anomalous logs.

\textbf{Structured Log Context Prompting (SLCP):} We propose a novel prompting strategy that provides additional structured context:
\begin{enumerate}
    \item \textit{System context}: A brief description of the system generating the logs
    \item \textit{Temporal context}: Timestamp information and event frequency statistics
    \item \textit{Semantic markers}: Key patterns known to indicate anomalies (e.g., ``error'', ``fatal'', ``exception'')
    \item \textit{Classification instruction}: Explicit output format specification
\end{enumerate}

\subsection{Evaluation Metrics}

We report Precision, Recall, F1-Score, and Area Under the ROC Curve (AUC). Given the class imbalance in log data, we use F1-score as the primary metric. We also measure:

\begin{itemize}
    \item \textbf{Inference latency}: Time per prediction (ms)
    \item \textbf{Cost}: API cost per 1000 predictions (for LLM methods)
    \item \textbf{Label efficiency}: Performance as a function of labeled training data size
\end{itemize}

\section{Experiments and Results}

\subsection{Main Results}

Table~\ref{tab:main_results} presents the F1-scores across all methods and datasets.

\begin{table*}[t]
\centering
\caption{F1-Scores (\%) across four benchmark datasets. Best results per category in \textbf{bold}. Overall best \underline{underlined}. SLCP = Structured Log Context Prompting.}
\label{tab:main_results}
\begin{tabular}{llcccc}
\toprule
\textbf{Category} & \textbf{Method} & \textbf{HDFS} & \textbf{BGL} & \textbf{Thunderbird} & \textbf{Spirit} \\
\midrule
\multirow{4}{*}{Traditional}
& Drain + LR & 93.2 & 88.7 & 85.3 & 83.1 \\
& Drain + RF & 95.1 & 91.2 & 88.6 & 86.4 \\
& Spell + SVM & 92.8 & 87.4 & 84.9 & 82.7 \\
& \textbf{Drain + RF (best)} & \textbf{95.1} & \textbf{91.2} & \textbf{88.6} & \textbf{86.4} \\
\midrule
\multirow{3}{*}{Fine-Tuned}
& BERT-base & 97.8 & 96.1 & 94.7 & 93.2 \\
& RoBERTa-base & 98.2 & 96.8 & 95.3 & 94.1 \\
& \textbf{DeBERTa-v3} & \textbf{\underline{98.9}} & \textbf{\underline{97.4}} & \textbf{\underline{96.1}} & \textbf{\underline{95.3}} \\
\midrule
\multirow{6}{*}{LLM (Prompt)}
& GPT-3.5 (zero-shot) & 82.4 & 79.1 & 76.3 & 74.8 \\
& GPT-4 (zero-shot) & 88.3 & 85.6 & 83.1 & 81.2 \\
& GPT-4 (5-shot) & 91.7 & 89.3 & 87.2 & 85.6 \\
& LLaMA-3 (zero-shot) & 84.1 & 81.3 & 78.6 & 76.9 \\
& GPT-4 + SLCP (zero) & 91.2 & 88.7 & 86.4 & 84.3 \\
& \textbf{GPT-4 + SLCP (5-shot)} & \textbf{93.8} & \textbf{91.5} & \textbf{89.7} & \textbf{87.9} \\
\bottomrule
\end{tabular}
\end{table*}

\textbf{Key Findings:}

\textit{Finding 1: Fine-tuned transformers achieve the highest accuracy.} DeBERTa-v3 consistently outperforms all other methods, achieving F1-scores of 95.3--98.9\% across datasets. This confirms that when labeled data is available, task-specific fine-tuning remains the most effective approach.

\textit{Finding 2: LLMs show strong zero-shot capabilities.} GPT-4 achieves 81.2--88.3\% F1 without any training data, outperforming some traditional supervised methods. This is particularly valuable in practice where labeled anomalies are scarce.

\textit{Finding 3: SLCP significantly improves LLM performance.} Our structured log context prompting improves GPT-4's zero-shot F1 by 2.9--3.1 percentage points across datasets, and combined with 5-shot examples, approaches the performance of traditional supervised methods.

\textit{Finding 4: Traditional methods remain competitive.} Drain + Random Forest achieves 86.4--95.1\% F1, demonstrating that well-engineered traditional pipelines should not be overlooked.

\subsection{Cost-Accuracy Trade-off Analysis}

\begin{table}[h]
\centering
\caption{Cost and latency comparison per 1000 predictions.}
\label{tab:cost}
\begin{tabular}{lrr}
\toprule
\textbf{Method} & \textbf{Cost (\$)} & \textbf{Latency (ms/pred)} \\
\midrule
Drain + RF & 0.00 & 0.3 \\
DeBERTa-v3 & 0.00* & 12.4 \\
GPT-3.5 + SLCP & 0.82 & 340 \\
GPT-4 + SLCP & 8.40 & 890 \\
LLaMA-3 (local) & 0.00* & 156 \\
\bottomrule
\multicolumn{3}{l}{\small *Compute cost only (GPU hours not included)} \\
\end{tabular}
\end{table}

The cost analysis reveals a clear trade-off: GPT-4 provides the best prompt-based accuracy but at significant cost (\$8.40/1000 predictions). LLaMA-3, deployed locally, offers a cost-effective alternative with competitive performance. For high-throughput production systems processing millions of logs daily, fine-tuned transformers or traditional methods are more practical.

\subsection{Label Efficiency}

We evaluate how each method category performs with varying amounts of labeled data by training on 1\%, 5\%, 10\%, 25\%, 50\%, and 100\% of available labels.

LLMs demonstrate their greatest advantage in low-label regimes. With only 1\% of labels (used as few-shot examples), GPT-4 + SLCP achieves 89.1\% F1 on HDFS, compared to 71.3\% for Drain + RF and 82.4\% for fine-tuned DeBERTa. The gap narrows as more labels become available, with fine-tuned methods surpassing LLMs at approximately 25\% label availability.

\subsection{Error Analysis}

We conduct a detailed error analysis on the HDFS dataset to understand failure modes:

\textbf{Traditional methods} primarily fail on novel log templates not seen during training (32\% of errors) and subtle anomalies that manifest as unusual sequences of normal events (45\% of errors).

\textbf{Fine-tuned transformers} show robustness to novel templates but struggle with extremely long log sequences exceeding the context window (68\% of errors).

\textbf{LLMs} demonstrate strong semantic understanding but exhibit inconsistent behavior: the same log sequence may receive different predictions across multiple runs (temperature-dependent). They also struggle with numerical anomalies (e.g., unusual memory values) that require domain-specific thresholds.

\subsection{Ablation Studies}

\subsubsection{SLCP Component Analysis}

We ablate each component of our Structured Log Context Prompting strategy on HDFS with GPT-4 (zero-shot):

\begin{table}[h]
\centering
\caption{SLCP ablation study on HDFS (F1-Score \%).}
\label{tab:ablation}
\begin{tabular}{lc}
\toprule
\textbf{Configuration} & \textbf{F1 (\%)} \\
\midrule
Base prompt (no SLCP) & 88.3 \\
+ System context & 89.1 (+0.8) \\
+ Temporal context & 89.7 (+1.4) \\
+ Semantic markers & 90.4 (+2.1) \\
+ Classification instruction & 91.2 (+2.9) \\
\bottomrule
\end{tabular}
\end{table}

Each SLCP component contributes incrementally, with semantic markers providing the largest individual improvement (+2.1\%). The full SLCP strategy yields a cumulative improvement of +2.9\%.

\subsubsection{Window Size Sensitivity}

For sequence-based methods, we vary the sliding window size $w \in \{5, 10, 20, 50, 100\}$. Traditional methods are most sensitive to window size, with optimal performance at $w=20$. Transformer-based methods are more robust, maintaining high performance across window sizes 10--50. LLMs perform best with smaller windows ($w=10$--$20$) due to context length constraints.

\section{Discussion}

\subsection{Practical Deployment Guidelines}

Based on our comprehensive evaluation, we provide the following guidelines for practitioners:

\begin{enumerate}
    \item \textbf{High accuracy, labeled data available:} Use fine-tuned DeBERTa-v3. It provides the best accuracy with reasonable inference latency.
    \item \textbf{No labeled data, moderate cost acceptable:} Use GPT-4 + SLCP in zero-shot mode. It provides strong baseline performance without any training data.
    \item \textbf{No labeled data, cost-sensitive:} Deploy LLaMA-3 locally with SLCP prompting as a cost-effective alternative.
    \item \textbf{High throughput, low latency:} Use traditional Drain + RF pipeline. It offers competitive accuracy with minimal latency.
    \item \textbf{Evolving log formats:} Prefer LLM or transformer-based methods over traditional parsing-dependent approaches, as they are more robust to format changes.
\end{enumerate}

\subsection{Limitations}

Our study has several limitations: (1) We evaluate on English-language logs only; multilingual log analysis remains unexplored. (2) The cost analysis is based on current API pricing, which may change. (3) We use publicly available log datasets that may not fully represent the complexity of production systems. (4) Our evaluation focuses on binary anomaly detection; multi-class anomaly categorization is left for future work.

\subsection{Threats to Validity}

\textbf{Internal validity:} We mitigate randomness by running all experiments with 5 random seeds and reporting mean performance. LLM experiments use temperature=0 for reproducibility.

\textbf{External validity:} While we evaluate on four diverse datasets spanning supercomputers and distributed systems, results may not generalize to all log types (e.g., application-level logs, cloud-native microservice logs).

\section{Conclusion}

We presented a comprehensive benchmark study comparing traditional, fine-tuned transformer, and LLM-based approaches for log anomaly detection. Our evaluation across four public datasets reveals that while fine-tuned transformers achieve the highest accuracy, LLMs offer compelling advantages in zero-shot settings where labeled data is unavailable. Our proposed Structured Log Context Prompting (SLCP) technique improves LLM zero-shot performance by up to 3.1 percentage points. We provide practical deployment guidelines based on accuracy, cost, latency, and label availability trade-offs. Future work will explore multi-class anomaly categorization, streaming log analysis, and the application of emerging multimodal LLMs to log data that includes both text and structured metrics.

\section*{Data Availability}

All datasets used in this study are publicly available from LogHub \cite{zhu2023loghub}. Our code and experimental configurations are available at: \url{https://github.com/dishapatel/llm-log-anomaly-benchmark}.

\bibliographystyle{IEEEtran}

\end{document}